%% file: main.tex
\documentclass[letterpaper, 10 pt, conference]{ieeeconf}  

\IEEEoverridecommandlockouts                              

\title{\LARGE \bf Timestamp-Supervised Action Segmentation with Graph Convolutional Networks}

\author{Hamza Khan~~~~~~~~~~Sanjay Haresh~~~~~~~~~~Awais Ahmed~~~~~~~~~~Shakeeb Siddiqui\\Andrey Konin~~~~~~~~~~M. Zeeshan Zia~~~~~~~~~~Quoc-Huy Tran
	\thanks{All authors are with Retrocausal, Inc., Redmond, WA 98052, USA.\newline
	Website: \url{www.retrocausal.ai}\newline
	Email: {\tt\small \{hamza,sanjay,awais,shakeeb,andrey,\newline
	zeeshan,huy\}@retrocausal.ai}}
}

\usepackage{graphicx}
\usepackage{amsmath}
\usepackage{amssymb}
\usepackage{booktabs}
\usepackage{soul}
\usepackage{hyperref}
\hypersetup{
    colorlinks=true,
    linkcolor=blue,
    urlcolor=magenta
    }
    
\usepackage{cite}
\usepackage{ctable}
\usepackage{array}
\usepackage{multirow}
\usepackage{float}
\usepackage{arydshln}
\usepackage{mwe}

\begin{document}

\maketitle
\thispagestyle{empty}
\pagestyle{empty}

\begin{abstract}
We introduce a novel approach for temporal activity segmentation with timestamp supervision. Our main contribution is a graph convolutional network, which is learned in an end-to-end manner to exploit both frame features and connections between neighboring frames to generate dense framewise labels from sparse timestamp labels. The generated dense framewise labels can then be used to train the segmentation model. In addition, we propose a framework for alternating learning of both the segmentation model and the graph convolutional model, which first initializes and then iteratively refines the learned models. Detailed experiments on four public datasets, including 50 Salads, GTEA, Breakfast, and Desktop Assembly, show that our method is superior to the multi-layer perceptron baseline, while performing on par with or better than the state of the art in temporal activity segmentation with timestamp supervision.
\end{abstract}

\input{Sections/introduction.tex}
\input{Sections/relatedwork.tex}
\input{Sections/method.tex}
\input{Sections/experiments.tex}
\input{Sections/conclusion.tex}

{\small
\bibliographystyle{IEEEtran}
\bibliography{references}
}

\end{document}

%% file: Sections/introduction.tex
\section{Introduction}
\label{sec:introduction}

Human activity understanding in videos has been an important research topic in the fields of robotics and computer vision, with various applications ranging from human-robot interaction, assisted living, healthcare, home automation to manufacturing~\cite{riek2017healthcare,iqbal2019human,konin2020retroactivity}. With the significant research efforts in the last decade, one can expect great results for the task of action recognition~\cite{tran2018closer,wang2018non}, where the input video is trimmed and captures a simple action. However, in many real-world applications~\cite{chao2018rethinking,gong2019memorizing}, we are often required to deal with untrimmed videos containing a complex activity. In this paper, we want to tackle one such problem, i.e., temporal activity segmentation, where given an input video capturing a complex activity, the task is to assign each frame of the video to one of the action/sub-activity classes.

The best performing methods for temporal activity segmentation are fully-supervised approaches~\cite{kuehne2016end, singh2016multi,richard2017weakly,farha2019ms, li2020ms}, where framewise action class labels are required during training. However, these framewise labels are difficult and costly to obtain. In contrast, unsupervised approaches~\cite{malmaud2015s, sener2015unsupervised,alayrac2016unsupervised,kukleva2019unsupervised, vidalmata2021joint,li2021action,kumar2021unsupervised}, which require little annotation, have been developed in the literature. Nevertheless, their performances are significantly worse than the above fully-supervised counterparts, which limits their practical applications. In order to balance between annotation efforts and segmentation accuracies, weakly-supervised approaches~\cite{huang2016connectionist, kuehne2017weakly, ding2018weakly,chang2019d3tw,richard2018action, fayyaz2020sct, li2020set,li2021temporal}, which leverage different forms of weak supervision, have attracted notable research interest. Here, we are particularly interested in timestamp supervision~\cite{li2021temporal} due to its superior accuracy as compared to other types of weak supervision~\cite{huang2016connectionist, kuehne2017weakly, ding2018weakly,chang2019d3tw,richard2018action, fayyaz2020sct, li2020set}.

\begin{figure}[t]
	\centering
		\includegraphics[width=0.9\linewidth, trim = 0mm 55mm 95mm 0mm, clip]{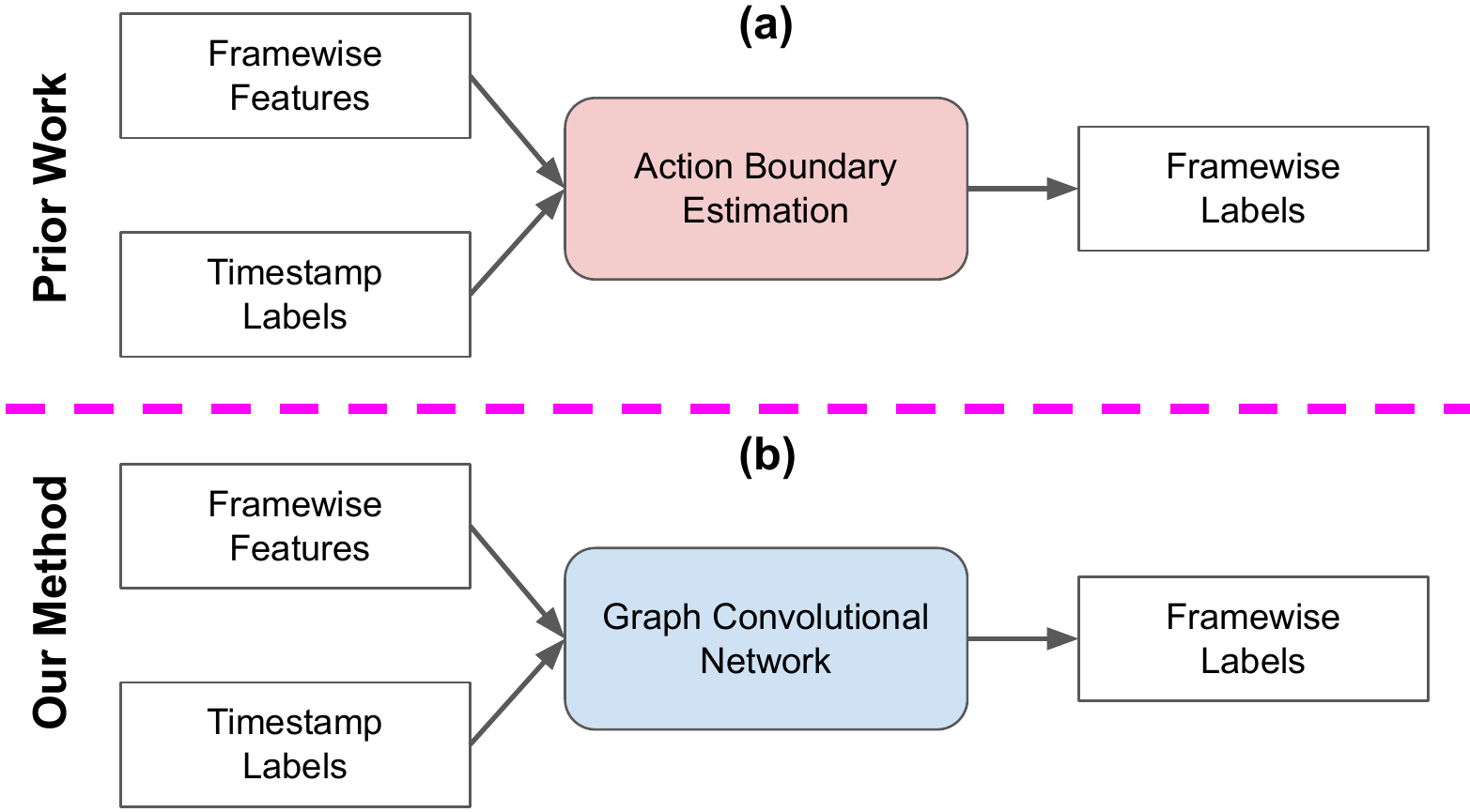}
	\caption{(a) Li et al.~\cite{li2021temporal} introduce a \emph{heuristic} action boundary estimation module which uses framewise features and sparse timestamp labels to generate dense framewise labels for training a segmentation model. In particular, it detects action boundaries by minimizing distances between frame features and action centers. (b) In contrast, we propose a \emph{learning} module based on graph convolutional network. More specifically, it leverages both frame features and connections between neighboring frames to convert sparse timestamp labels to dense framewise labels.}
	\label{fig:teaser}
\end{figure}

In timestamp-supervised setup~\cite{moltisanti2019action,li2021temporal}, only one (random) frame is annotated for each action segment in a training video. Naively training the segmentation model with just these sparse timestamp labels leads to suboptimal results since the major (unlabeled) part of the video is not used effectively. In this work, we propose to learn a graph convolutional network to convert sparse timestamp labels to dense framewise labels for training the segmentation model (see Fig.~\ref{fig:teaser}(b)). Our work is motivated by~\cite{kipf2016semi}, where a graph convolutional network is utilized for effectively propagating labels from only a few labeled nodes to the remaining unlabeled nodes. In our work, the graph convolutional network is learned in an end-to-end manner to exploit not only frame features but also connections between neighboring frames. This is in contrast to the heuristic action boundary estimation model in~\cite{li2021temporal}, which detects action boundaries by minimizing distances between frame features and action centers to generate dense framewise labels (see Fig.~\ref{fig:teaser}(a)). In addition, to effectively learn both the segmentation model and the graph convolutional model, we introduce an alternating learning framework, which first initializes the learned models from scratch and then iteratively improves them.

In summary, our contributions include:
\begin{itemize}
    \item We present a novel method for activity segmentation with timestamp supervision. Our main idea is to learn a graph convolutional network, which exploits both frame features and connections between neighboring frames to transform sparse timestamp labels to dense framewise labels. Moreover, we develop a framework for alternating learning of both the segmentation model and the graph convolutional network.
    \item Extensive evaluations demonstrate that our method outperforms the multi-layer perceptron baseline, and performs on par with or better than the state of the art in timestamp-supervised activity segmentation on 50 Salads, GTEA, Breakfast, and Desktop Assembly datasets.
\end{itemize}

%% file: Sections/relatedwork.tex
\section{Related Work}
\label{sec:relatedwork}

Below we summarize related works in temporal activity segmentation and graph convolutional networks with a focus on applications in video understanding problems.

\noindent \textbf{Fully-Supervised Activity Segmentation.} Temporal activity segmentation plays an important role in understanding human activities~\cite{pirsiavash2014parsing,caba2015activitynet,damen2018scaling}. Fully-supervised methods reason about long-range dependencies in videos, which are typically tackled by using a two-stage approach of feature extraction and temporal reasoning with an HMM or RNN~\cite{kuehne2016end, singh2016multi, richard2017weakly}. Recent approaches~\cite{farha2019ms, li2020ms} show that single-stage temporal convolutions are capable of modeling long-range dependencies effectively. Despite satisfactory results, the above approaches need dense framewise labels for fully-supervised training. Our method, on the other hand, requires weak supervision in the form of sparse timestamp labels.

\noindent \textbf{Weakly-Supervised Activity Segmentation.} Many works have focused on reducing the amount of annotations by using transcript supervision~\cite{huang2016connectionist, kuehne2017weakly, ding2018weakly,chang2019d3tw}, i.e., the order of actions occurring in a given video. They mostly rely on aligning frames with transcripts by using different alignment techniques such as Dynamic Time Warping~\cite{chang2019d3tw} and Viterbi~\cite{richard2018neuralnetwork}. Other approaches have also experimented with relaxing the ordering assumption by using only a set of actions for supervision, namely set supervision~\cite{richard2018action, fayyaz2020sct, li2020set}. Although these algorithms are efficient in terms of significantly less annotation requirement, they perform notably worse than fully-supervised approaches. To alleviate that, Li et al.~\cite{li2021temporal} propose to use timestamp supervision, i.e., for each action segment in a training video, only one (random) frame is annotated. Our method also uses timestamp supervision, however, instead of heuristic action boundary estimation in~\cite{li2021temporal}, we learn a graph convolutional network to generate dense framewise labels.

\noindent \textbf{Unsupervised Activity Segmentation.} Unsupervised activity segmentation has received considerable research interest due to little annotation requirement. Early methods~\cite{malmaud2015s, sener2015unsupervised,alayrac2016unsupervised} leverage narration cues from accompanying scripts to segment videos. Recent methods~\cite{kukleva2019unsupervised, vidalmata2021joint} use pretext tasks for learning frame representations before clustering them by using $K$-means clustering to form action clusters. Furthermore, Li et al.~\cite{li2021action} exploit action-level cues to boost the performance. More recently, Kumar et al.~\cite{kumar2021unsupervised} perform frame representation learning and clustering in a joint framework via temporal optimal transport. Despite little annotation requirement, the above methods perform significantly worse than fully-supervised or weakly-supervised approaches.

\noindent \textbf{Graph Convolutional Networks.} Graph convolutional networks, e.g., \cite{kipf2016semi}, have been shown to work more effectively on graph structured data in label-scarce settings than classical neural network architectures. They have also been applied recently to various video understanding problems such as action recognition~\cite{herzig2019spatio}, object-object reasoning~\cite{haresh2020towards}, and action localization~\cite{zeng2019graph}. While the above works have focused on using graph convolutional networks for video understanding problems in fully-supervised settings, we leverage graph convolutional networks to generate dense framewise labels for activity segmentation in semi-supervised settings.

%% file: Sections/method.tex
\section{Our Approach}
\label{sec:method}

\begin{figure*}[t]
	\centering
		\includegraphics[width=0.9\linewidth, trim = 0mm 40mm 15mm 0mm, clip]{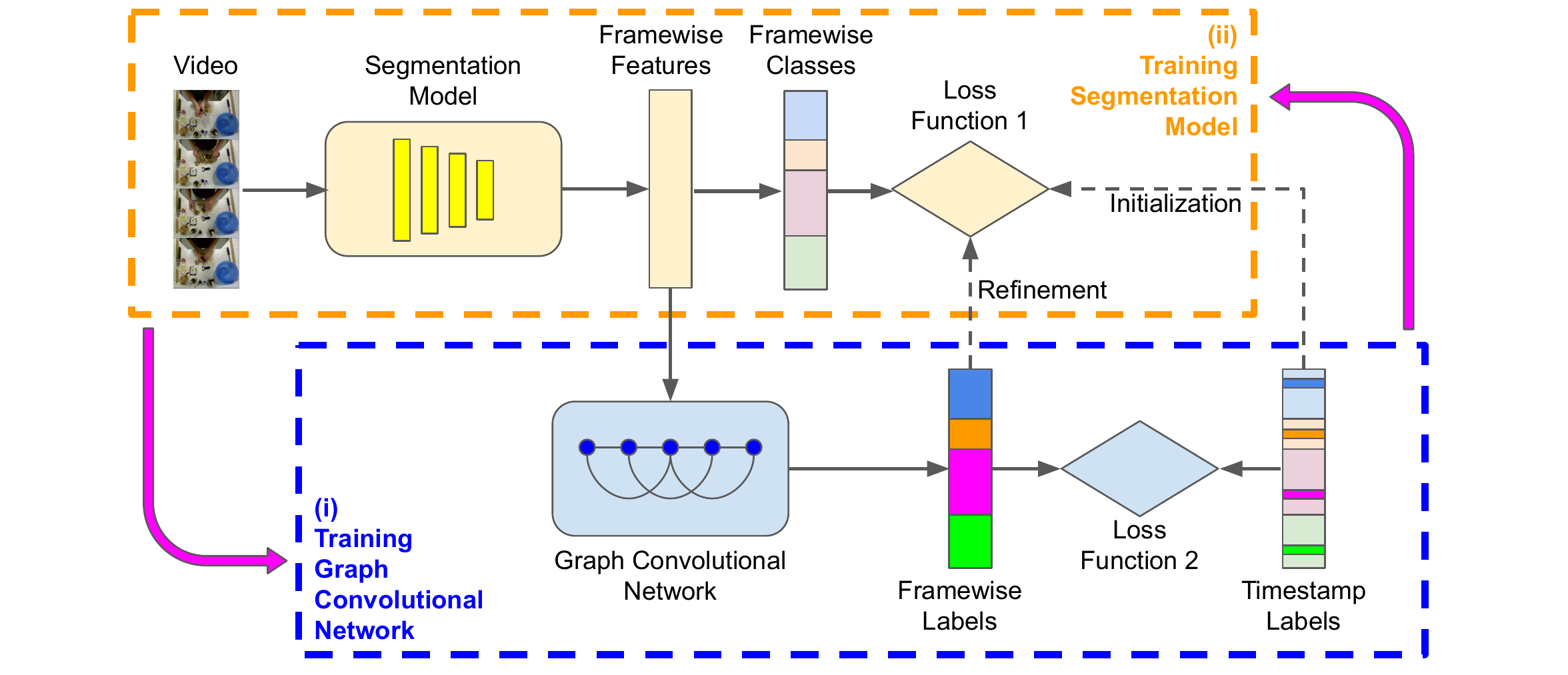}
	\caption{Our alternating learning framework is shown above. During training, we first initialize the segmentation model by training it with sparse timestamp labels (namely, the \emph{initialization} stage). After that, we alternate between (i) training the graph convolutional network, which takes framewise features from the segmentation model as input, uses sparse timestamp labels for supervision, and generates dense framewise labels as output, and (ii) training the segmentation model, which uses dense framewise labels from the graph convolutional network for supervision (namely, the \emph{refinement} stage). During testing, the graph convolutional network is discarded while the segmentation model is employed to provide segmentation results.}
	\label{fig:overview}
\end{figure*}

We present, in this section, the details of our approach. We first describe the task of temporal activity segmentation and timestamp supervision in Sec.~\ref{sec:activity_segmentation}. Next, the proposed graph convolutional network module for generating dense framewise labels is introduced in Sec.~\ref{sec:gcn}. Lastly, we propose a framework for alternating learning of the segmentation model and the graph convolutional network in Sec.~\ref{sec:alternating_learning}.

\subsection{Activity Segmentation with Timestamp Supervision}
\label{sec:activity_segmentation}

Temporal activity segmentation aims to associate each frame of an input video capturing a complex activity with one of the action/sub-activity classes. In particular, let us denote the input video as $X = [x_1, x_2, \dots, x_T]$, where $T$ is the number of frames in $X$. The task is to obtain the action class for each frame of $X$, i.e., $A = [a_1, a_2, \dots, a_T]$ with $a_i$ representing the action class for frame $x_i$. For fully-supervised methods~\cite{farha2019ms,li2020ms,wang2020boundary,ishikawa2021alleviating}, a number of videos $\mathcal{X} = \{X\}$ with corresponding framewise labels $\mathcal{A} = \{A\}$ are given for training. These methods focus on designing an effective segmentation model  to extract useful appearance and temporal cues in the videos and often produce satisfactory results. However, framewise annotations for all training videos are generally difficult and prohibitively expensive to obtain.

In this work, we are interested in the timestamp-supervised setting~\cite{moltisanti2019action,li2021temporal}. Specifically, for each action segment in a training video, only one (random) frame is labeled. Assuming a training video $X$ has $N$ action segments (in general, $N \ll T$), we can denote the timestamp labels for $X$ as $A_{TS} = [a_{t_1}, a_{t_1}, \dots, a_{t_N}]$, where frame $t_i$ belongs to the $i$-th action segment in $X$. According to~\cite{ma2020sf}, timestamp-supervised setting requires a significantly less (i.e., one sixth) labeling duration as compared to the above fully-supervised setting. Nevertheless, simply using the sparse timestamp labels for training the segmentation model (i.e., applying the classification loss with these sparse timestamp labels) leads to inferior results since the remaining (unlabeled) frames are not utilized effectively. To address that, we propose, in the next section, a graph convolutional network module for generating dense framewise labels from sparse timestamp labels so that all frames can be employed for training.

\subsection{Graph Convolutional Network for Label Generation}
\label{sec:gcn}

Let us represent each video as a graph, where frames (along with their features) are nodes and there exist edges between neighboring frames (within a temporal window). Given only a few nodes with labels (i.e., sparse timestamp labels), we want to classify the remaining nodes in the graph. This problem, namely graph-based semi-supervised learning, is typically solved by minimizing the following loss function:
\begin{align}
    \mathcal{L} = \mathcal{L}_{class} + \lambda \mathcal{L}_{reg},
\end{align}
where $\mathcal{L}_{class}$ denotes the classification loss on the labeled nodes, $\mathcal{L}_{reg}$ represents the \emph{explicit} graph-based regularization term~\cite{zhu2003semi,zhou2003learning,belkin2006manifold,weston2012deep}, and $\lambda$ is the balancing parameter. 

Inspired by~\cite{kipf2016semi}, we encode the graph structure directly by using a graph neural network $f(\textbf{X},\textbf{A})$. Here, $\textbf{X}$ denotes the node features, \textbf{A} represents the adjacency matrix, and $f$ is a graph neural network. That enables us to train $f$ with only $\mathcal{L}_{class}$ since $\mathcal{L}_{reg}$ is \emph{implicitly} embedded in $f$. Depending $f$ on both the node features and the adjacency matrix allows gradient information to be distributed from the labeled nodes to the unlabeled ones. Thus, the model is able to learn effective representations for both labeled and unlabeled nodes. In particular, we use a graph convolutional network with the below  propagation rule:
\begin{align}
    \textbf{H}_{l+1} = \sigma \left( \Tilde{\textbf{D}}^{-\frac{1}{2}} \Tilde{\textbf{A}} \Tilde{\textbf{D}}^{-\frac{1}{2}} \textbf{H}_{l} \textbf{W}_{l} \right).
\end{align}
Here, $\Tilde{\textbf{A}} = \textbf{A} + \textbf{I}$ is the adjacency matrix with added self-connections (represented by the identity matrix $\textbf{I}$), while $\Tilde{\textbf{D}}$ is the degree matrix of $\Tilde{\textbf{A}}$. Next, $\textbf{W}_{l}$, $\sigma$, and $\textbf{H}_{l}$ are the weight matrix for layer $l$, the activation function, and the activation matrix for layer $l$ respectively. As demonstrated in~\cite{kipf2016semi}, the above graph convolutional network can be derived from a first-order approximation of spectral graph convolutions~\cite{hammond2011wavelets}.

As we empirically show in Sec.~\ref{sec:abl_graph_struct}, using weighted graphs yields better results than using binary graphs. For weighted graphs, we define the edge weight between nodes $i$ and $j$ as the cosine similarity between corresponding features $\textbf{x}_i$ and $\textbf{x}_j$, which is written as:
\begin{align}
    \textbf{A}_{ij} = \frac{\textbf{x}_i \cdot \textbf{x}_j}{\| \textbf{x}_i\| \| \textbf{x}_j\|},
    \label{eq:edge_weight}
\end{align}
where $\cdot$ denotes the dot product.

\subsection{Alternating Learning Framework}
\label{sec:alternating_learning}

In the following, we present our alternating learning framework for training the segmentation model and the graph convolutional network. In particular, we divide the training into two stages, i.e., the initialization stage and the refinement stage. During the \emph{initialization} stage, we train only the segmentation model with sparse timestamp labels for $\epsilon$ epochs ($\gamma = 30$). Next, in the \emph{refinement} stage, we perform $\gamma$ iterations of alternating learning ($\epsilon = 20$). For each iteration, we (i) first train the graph convolutional network for 300 epochs, where framewise features from the segmentation model are input, sparse timestamp labels are supervision signals, and dense framewise labels are output, and (ii) then train the segmentation model for 3 epochs, where dense framewise labels from the graph convolutional network are supervision signals. At testing, the graph convolutional network is discarded, and the segmentation model is used to produce the segmentation result. Fig.~\ref{fig:overview} summarizes our alternating learning framework. As we will show later in Sec.~\ref{sec:abl_alternating_learning}, our alternating learning framework  for training the segmentation model and the graph convolutional network outperforms the joint learning counterpart.

We use the conventional combination of classification loss and smoothing loss~\cite{farha2019ms,li2020ms,wang2020boundary,ishikawa2021alleviating} for training the segmentation model and graph convolutional network. In addition, we include the confidence loss~\cite{li2021temporal} in the training of the segmentation model to boost its performance. We summarize those losses below.

\noindent \textbf{Classification Loss.} We apply the cross-entropy loss between the predicted probabilities and the (generated) action labels as:
\begin{align}
    \mathcal{L}_{class} = \frac{1}{T} \sum_{t} -\log \Tilde{y}_{t,a},
\end{align}
where $T$ is the number of frames in the video and $\Tilde{y}_{t,a}$ is the predicted probability that frame $x_t$ is assigned to action class $a$.

\noindent \textbf{Smoothing Loss.} We employ the smoothing loss to tackle the problem of over-segmentation as:
\begin{align}
    \mathcal{L}_{smooth} = \frac{1}{TC} \sum_{t,a} \Tilde{\Delta}^2_{t,a},\\
    \Tilde{\Delta}_{t,a} = \begin{cases}
        \Delta_{t,a}, & \Delta_{t,a} \leq \tau\\
        \tau, & \Delta_{t,a} > \tau\\
    \end{cases},\\
    \Delta_{t,a} = \left| \log \Tilde{y}_{t,a} - \log \Tilde{y}_{t-1,a} \right|,
\end{align}
where $C$ is the number of action classes in the activity and $\tau = 4$ is the thresholding parameter.

\noindent \textbf{Confidence Loss.} We adopt the confidence loss from~\cite{li2021temporal} to encourage the predicted probabilities to monotonically decrease as the distance to the timestamps increases as:
\begin{align}
    \mathcal{L}_{conf} = \frac{1}{T'} \sum_{a_{t_i} \in A_{TS}} \left( \sum_{t = t_{i-1}}^{t_{i+1}} \delta_{a_{t_i},t} \right),\\
    \delta_{a_{t_i},t} = \begin{cases}
        \max (0, \log \Tilde{y}_{t,a_{t_i}} - \log \Tilde{y}_{t-1,a_{t_i}}), & t \leq t_i\\
        \max (0, \log \Tilde{y}_{t-1,a_{t_i}} - \log \Tilde{y}_{t,a_{t_i}}), & t > t_i\\
    \end{cases},
\end{align}
where $t_i$ and $a_{t_i}$ are the $i$-th timestamp and its corresponding action label, $\Tilde{y}_{t,a_{t_i}}$ is the predicted probability that frame $x_t$ is assigned to action class $a_{t_i}$, and $T' = 2(t_N-t_1)$ is the number of frames contributing to the loss.

\noindent \textbf{Final Losses.} The final losses $\mathcal{L}_{seg}$ and $\mathcal{L}_{graph}$ respectively for training the segmentation model and the graph convolutional network are written as follows:
\begin{align}
    \mathcal{L}_{seg} = \mathcal{L}_{class} + \alpha \mathcal{L}_{smooth} + \beta \mathcal{L}_{conf},\\
    \mathcal{L}_{graph} = \mathcal{L}_{class} + \alpha \mathcal{L}_{smooth}.
\end{align}
 Here, $\alpha = 0.15$ and $\beta = 0.075$ are balancing parameters. We have tried adding the confidence loss $\mathcal{L}_{conf}$ to the final loss $\mathcal{L}_{graph}$ for training the graph convolutional network but did not get any performance gain.

%% file: Sections/experiments.tex
\section{Experiments}
\label{sec:experiments}

In this section, we benchmark our approach for timestamp-supervised activity segmentation on various datasets, including 50 Salads, GTEA, Breakfast, and Desktop Assembly.

\noindent \textbf{Implementation Details.} For a fair comparison, we follow~\cite{li2021temporal} to adopt the multi-stage temporal convolutional network of~\cite{farha2019ms} as our segmentation model. The I3D features~\cite{carreira2017quo} used as input in~\cite{li2021temporal} are also used as input to our segmentation model. In addition, we implement a two-layer graph convolutional network for label generation. The first layer maps the input features ($64$-dimensional vectors) to $32$-dimensional vectors, which are subsequently passed through ReLU activation, the second layer, and lastly softmax classification. To construct the graph from an input video, we consider frames along with their features as nodes and connect each frame with its preceding $15$ frames and succeeding $15$ frames (yielding a temporal window size of $31$) to form pairwise edges. The input features to the graph convolutional network are the output of the penultimate layer of the segmentation model and are $64$-dimensional vectors. The segmentation model and the graph convolutional network are learned via backpropagation respectively through the losses in Sec.~\ref{sec:alternating_learning}. We implement our approach in PyTorch~\cite{paszke2017automatic}. We use the ADAM optimizer~\cite{kingma2014adam} and a batch size of $8$. The learning rate is set to $0.0005$ and $0.01$ respectively for the segmentation model and the graph convolutional network. We use a weight decay of $0.0005$ only for the graph convolutional network. Our experiments are conducted with an Nvidia V100 GPU on Microsoft Azure.

\noindent \textbf{Competing Methods.} We compare our approach (namely ``\emph{GCN}'', short for \emph{G}raph \emph{C}onvolutional \emph{N}etwork) against the state-of-the-art method of\cite{li2021temporal}\footnote{We use the original code provided by the authors at\\ \url{https://github.com/ZheLi2020/TimestampActionSeg}} (namely ``\emph{ABE}'', short for \emph{A}ction \emph{B}oundary \emph{E}stimation) for timestamp-supervised activity segmentation. Also, we add a fully-supervised baseline (namely ``\emph{Baseline}''), which has the same architecture as our segmentation model but is trained with ground truth dense framewise labels for full supervision. Lastly, we include the results of recent fully-supervised methods~\cite{farha2019ms,li2020ms,wang2020boundary,ishikawa2021alleviating}, transcript-supervised methods~\cite{huang2016connectionist,richard2017weakly,ding2018weakly,richard2018neuralnetwork,chang2019d3tw,li2019weakly,souri2021fast}, and set-supervised methods~\cite{richard2018action,li2020set,fayyaz2020sct}.

\noindent \textbf{Datasets.} We test the performance on four public datasets, namely 50 Salads~\cite{stein2013combining}, Breakfast~\cite{Kuehne_2014_CVPR}, GTEA~\cite{fathi2011learning}, and Desktop Assembly~\cite{kumar2021unsupervised}. We summarize the datasets below:
\begin{itemize}
    \item \emph{50 Salads}: The dataset consists of 50 videos with actors preparing different kinds of salads. It contains 0.6M frames annotated with one of the 17 action classes, and the average video duration varies from 5 to 10 minutes.
    \item \emph{Breakfast}: The dataset includes 1712 videos with actors preparing various types of breakfast. It consists of 3.6M frames annotated with one of the 48 action classes, and the average video duration ranges from a few seconds to several minutes.
    \item \emph{GTEA}: The dataset consists of 28 videos with actors performing various kinds of daily activities. It contains 32K frames annotated with one of the 11 action classes, and the average video duration is 1 minute.
    \item \emph{Desktop Assembly}: The dataset includes 76 videos with actors performing the desktop assembly activity. It consists of 59K frames annotated with one of the 22 action classes, and the average video duration is 1.5 minutes.
\end{itemize}
We use the same timestamp labels for our method and ABE.

\noindent \textbf{Metrics.} We follow~\cite{li2021temporal} to report the results on five metrics, including framewise accuracy (Acc), edit distance (Edit), and F1 scores with overlapping thresholds of 10\%, 25\%, and 50\%. To reduce the impact of randomness, we use $K$-fold cross validation with $K = 5$ for 50 Salads and $K = 4$ for GTEA, Breakfast, and Desktop Assembly. Furthermore, for timestamp-supervised methods, including ours and ABE, we run each 3 times with 3 random seeds and report the \emph{average} results over the 3 runs. We would like to note that this setup is different from the setup in~\cite{li2021temporal}, where the results are reported over a single run. For other types of methods, we simply obtain their results from~\cite{li2021temporal}.

\subsection{Ablation Study on Graph Construction}
\label{sec:abl_graph_struct}

\input{Tables/abl_graph_struct}

We first conduct some ablation study experiments to understand the impacts of graph structures on our approach. In particular, we evaluate the performance of our method when using different types of graphs, i.e., binary graphs and weighted graphs, as well as various temporal window sizes, i.e., $3$, $7$, $17$, and $31$. For weighted graphs, the edge weight is defined as in Eq.~\ref{eq:edge_weight}. Tab.~\ref{tab:abl_graph_struct} presents the ablation study results on graph construction on the 50 Salads and GTEA datasets. From the results, temporal window size of $3$ performs the worst, while larger temporal window sizes, which are able to capture longer-range dependencies despite having higher computational costs, often yield better results. Next, temporal window size of $31$ has the best performance on both 50 Salads and GTEA datasets. Further, we have tried with larger temporal window sizes than $31$, however, they yield little performance gains while having notably higher computational costs. In addition, by leveraging additional information (i.e., similarity between frame features), weighted graphs usually outperform binary graphs on both 50 Salads and GTEA datasets. For all the remaining experiments in this section, we will use temporal window size of $31$ and weighted graphs for our method.

\subsection{Ablation Study on MLP vs. GCN}
\label{sec:mlp_ablation}

\input{Tables/mlp_ablation}

As mentioned in Sec.~\ref{sec:gcn}, the graph convolutional network embeds the graph-based regularization term $\mathcal{L}_{reg}$ implicitly, which helps it exploit cues available in the graph (i.e., frame features and connections between neighboring frames). In this section, we study the benefit of leveraging these cues via a graph convolutional network as compared to using a plain multi-layer perceptron  network. Tab.~\ref{tab:mlp_ablation} shows the results on the 50 Salads and GTEA datasets when using a graph convolutional network (GCN) and a multi-layer perceptron network (MLP) for label generation. Note that they have the \emph{same} set of hyperparameters, e.g., number of layers, number of output channels, etc. As we can see from Tab.~\ref{tab:mlp_ablation}, GCN outperforms MLP across all metrics on both datasets, which confirms the advantage of using GCN over MLP for label generation.

\subsection{Ablation Study on Alternating Learning vs. Joint Learning}
\label{sec:abl_alternating_learning}

\input{Tables/abl_alternating_learning}

Here, we conduct experiments to support our choice of alternating learning over joint learning of the segmentation model and the graph convolutional network. In particular, we compare the performance of joint learning (i.e., without the initialization stage or $\{\gamma,\epsilon\}$ = $\{0,50\}$) with two versions of alternating learning (i.e., $\{\gamma,\epsilon\}$ = $\{10,40\}$ and $\{\gamma,\epsilon\}$ = $\{30,20\}$). Moreover, we compare with the naive learning of the segmentation model using only sparse timestamp labels (i.e., without the refinement stage or $\{\gamma,\epsilon\}$ = $\{50,0\}$). The results on the 50 Salads and GTEA datasets are shown in Tab.~\ref{tab:abl_alternating_learning}. It is evident from the results that alternating learning outperforms both joint learning and naive learning by large margins. Next, the bad performance of joint learning is likely because as we train both models from scratch, during the first few iterations, the features extracted by the segmentation model and hence the framewise labels predicted by the graph convolutional network are not meaningful. As a result, the joint learning framework is unstable and yields worse results than the alternating learning one. Lastly, the worse performance of naive learning is likely because unlabeled frames in the video is not utilized effectively. In the following experiments, we will use alternating learning with $\{\gamma,\epsilon\}$ = $\{30,20\}$ for our method.

\subsection{Comparison on 50 Salads}
\label{sec:50sld_diff_suprvsn}

\input{Tables/50sld_diff_suprvsn}

\begin{figure}[t]
	\centering
		\includegraphics[width=0.9\linewidth, trim = 0mm 0mm 0mm 0mm, clip]{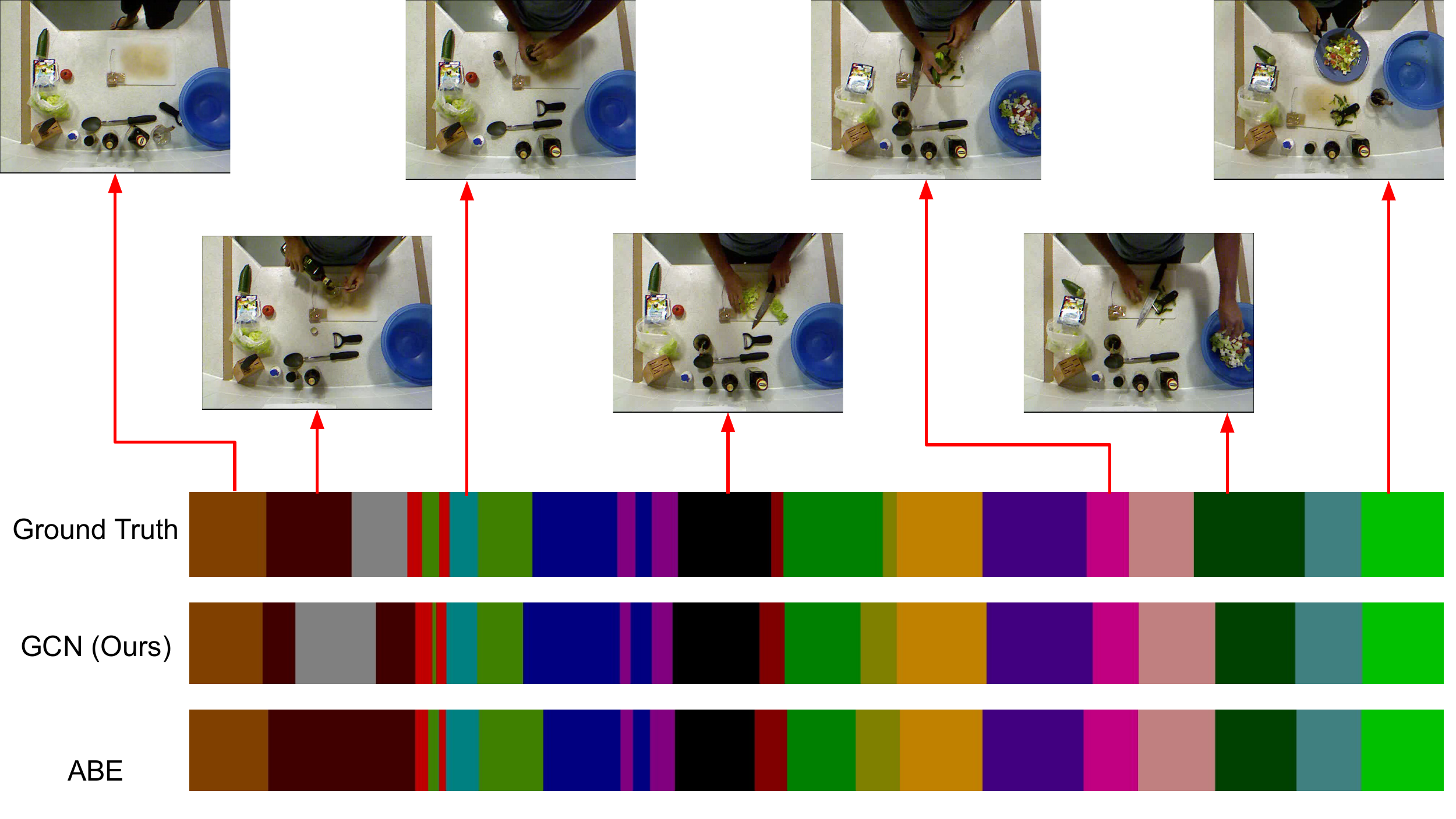}
	\caption{Segmentation results on a 50 Salads video.}
	\label{fig:50salads}
\end{figure}

We now compare the performance of our approach against that of the state-of-the-art timestamp-supervised method of~\cite{li2021temporal} (ABE), on the 50 Salads dataset. In addition, we include the results of fully-supervised methods~\cite{farha2019ms,li2020ms,wang2020boundary,ishikawa2021alleviating} and transcript-supervised methods~\cite{richard2017weakly,richard2018neuralnetwork,li2019weakly}. Tab.~\ref{tab:50salads_diff_supervision} presents the results. As we see from Tab.~\ref{tab:50salads_diff_supervision}, our approach outperforms ABE on F1@10, F1@25, F1@50, and Edit, while performing on par with ABE on Acc. Lower F1 and Edit indicate that ABE is more likely to suffer from over-segmentation, since over-segmented results may have high Acc but yield low F1 and Edit. In addition, timestamp-supervised methods produce better results than transcript-supervised ones. Fig.~\ref{fig:50salads} visualizes some qualitative results of our approach and ABE, where our result is closer to the ground truth. Please see also our supplementary video\footnote{Our supplementary video is available at\\ \url{https://youtu.be/tvV3soPMTIo}}.

\subsection{Comparison on GTEA}
\label{sec:gtea_diff_suprvsn}

\input{Tables/gtea_diff_suprvsn}

\begin{figure}[t]
	\centering
		\includegraphics[width=0.9\linewidth, trim = 0mm 0mm 0mm 0mm, clip]{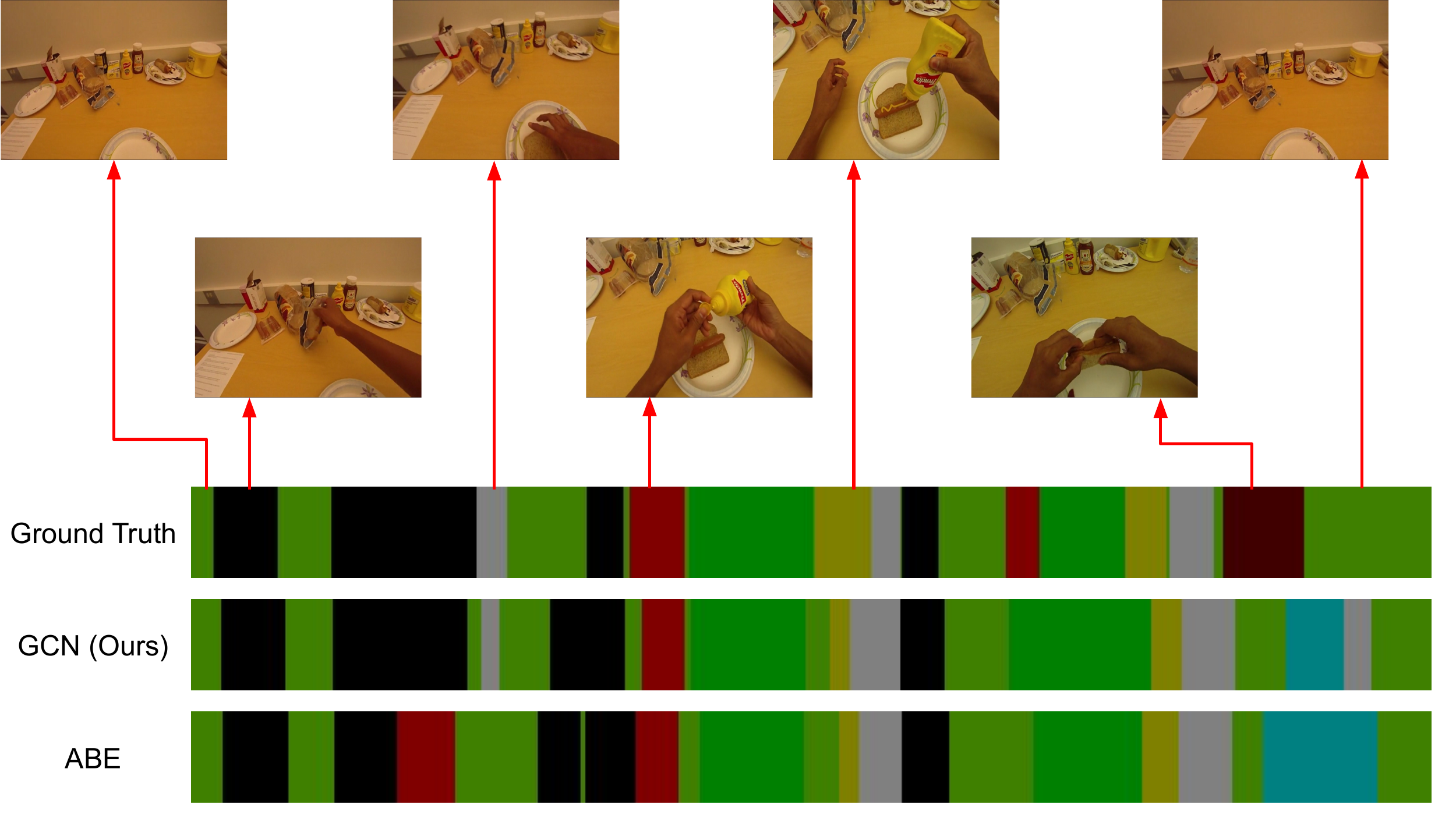}
	\caption{Segmentation results on a GTEA video.}
	\label{fig:gtea}
\end{figure}

Tab.~\ref{tab:gtea_diff_supervision} presents the results of timestamp-supervised methods, i.e., ours and ABE~\cite{li2021temporal}, and fully-supervised methods~\cite{farha2019ms,li2020ms,wang2020boundary,ishikawa2021alleviating} on the GTEA dataset. Similar to the results on 50 Salads, our approach achieves higher F1@10, F1@25, F1@50, and Edit, and lower Acc than ABE on GTEA. This also shows our method is less suffering from over-segmentation as compared to ABE. In addition, some qualitative results of our method and ABE are plotted in Fig.~\ref{fig:gtea}, where our result is closer to the ground truth.

\subsection{Comparison on Breakfast}
\label{sec:breakfast_diff_suprvsn}

\input{Tables/breakfast_diff_suprvsn}

\begin{figure}[t]
	\centering
		\includegraphics[width=0.9\linewidth, trim = 0mm 0mm 0mm 0mm, clip]{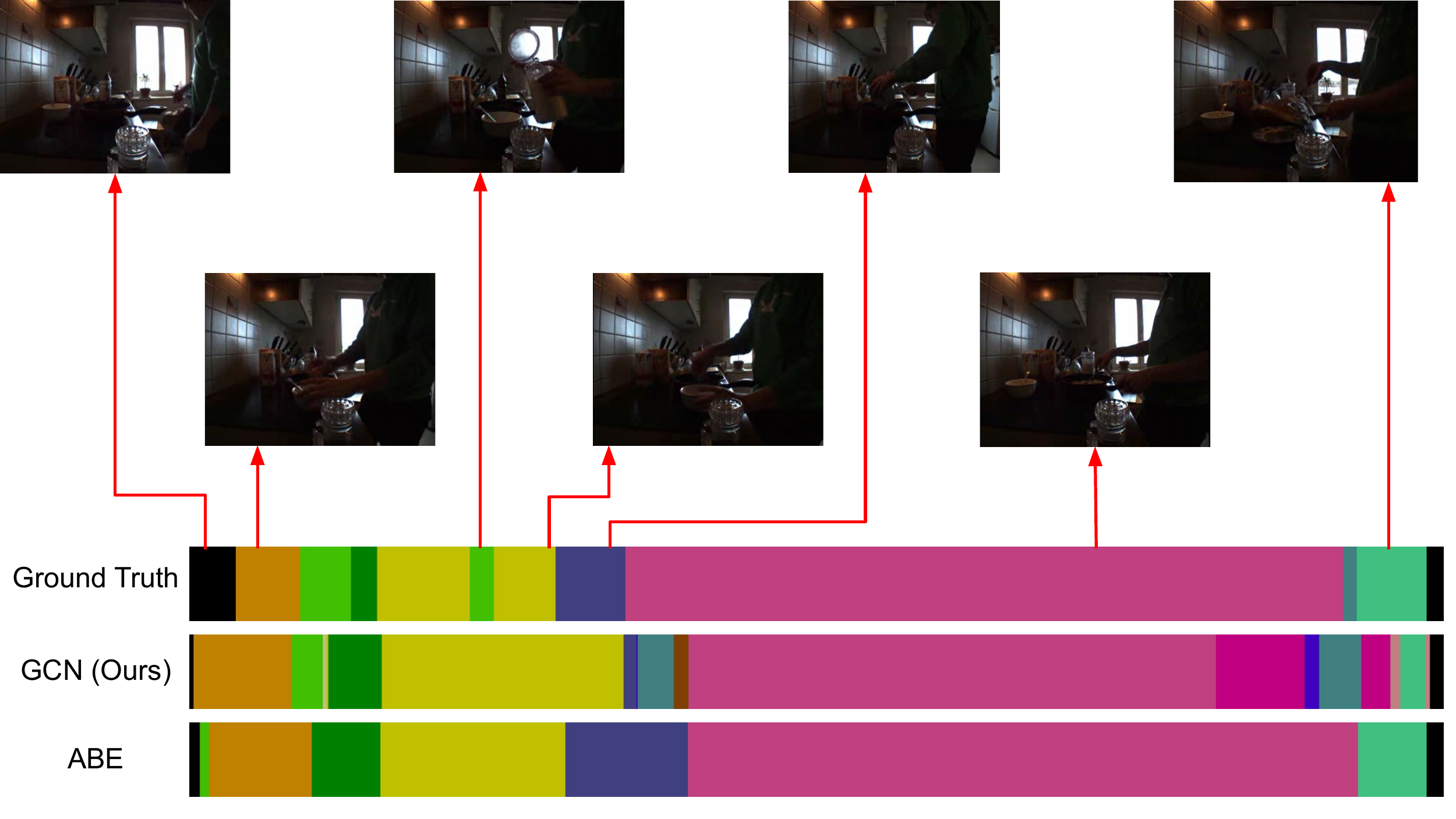}
	\caption{Segmentation results on a Breakfast video.}
	\label{fig:breakfast}
\end{figure}

The results of methods with different forms of supervision, including full supervision~\cite{farha2019ms,li2020ms,wang2020boundary,ishikawa2021alleviating}, timestamp supervision (ours and ABE~\cite{li2021temporal}), transcript supervision~\cite{huang2016connectionist,richard2017weakly,ding2018weakly,richard2018neuralnetwork,chang2019d3tw,li2019weakly,souri2021fast}, and set supervision~\cite{richard2018action,li2020set,fayyaz2020sct}, on the Breakfast dataset are shown in Tab.~\ref{tab:breakfast_diff_supervision}. From Tab.~\ref{tab:breakfast_diff_supervision}, our approach obtains similar results as ABE (i.e., our method has higher F1@10, F1@25, and F1@50, and lower Edit and Acc). In addition, timestamp-supervised methods yield better results than other weakly-supervised ones. Fig.~\ref{fig:breakfast} plots some qualitative results of our approach and ABE, where the two methods perform similarly.

\subsection{Comparison on Desktop Assembly}
\label{sec:desktopassembly_diff_suprvsn}

\input{Tables/desktopassembly_diff_suprvsn}

\begin{figure}[t]
	\centering
		\includegraphics[width=0.9\linewidth, trim = 0mm 0mm 0mm 0mm, clip]{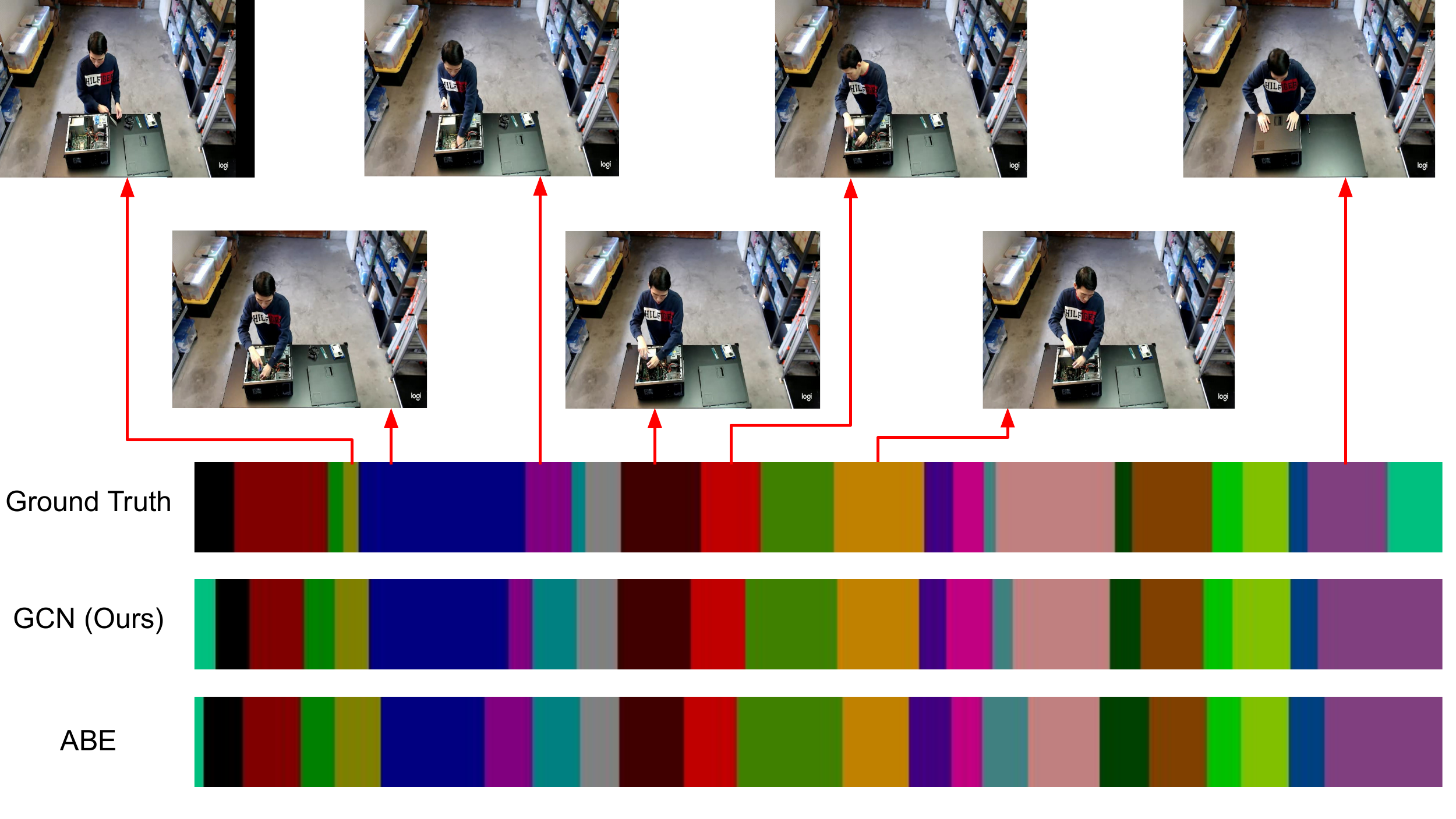}
	\caption{Segmentation results on a Desktop Assembly video.}
	\label{fig:desktopassembly}
\end{figure}

Tab.~\ref{tab:desktopassembly_diff_supervision} presents the results of timestamp-supervised methods, namely ours and ABE~\cite{li2021temporal}, and fully-supervised methods, including the state-of-the-art method of~\cite{ishikawa2021alleviating}, on the Desktop Assembly dataset. From the results, our approach outperforms ABE on F1@10, F1@25, F1@50, and Acc by substantial margins, while performing on par with ABE on Edit. In addition, Fig.~\ref{fig:desktopassembly} visualizes some qualitative results of our approach and ABE, where our result is closer to the ground truth.

%% file: Tables/abl_graph_struct.tex
\begin{table}[t]
\centering
\renewcommand{\arraystretch}{1.5}
\begin{tabular}{c|c|c|c c c c c}
\toprule
& Edge & Window & F1@10 &  F1@25 & F1@50 & Edit & Acc\\
\midrule

\multirow{8}{*}{\scriptsize{\textbf{F}}}

& \multirow{4}{*}{\rotatebox[origin=c]{90}{\scriptsize{Binary}}}

& 3  & 73.7 & 70.5 & 59.1 & 65.7 & 73.6\\

& & 7  & 73.9 & 71.0 & 59.2 & 66.4 & 74.1\\
 
& & 17 & 74.2 & 71.1 & 58.9 & 66.5 & 74.7\\

& & 31  & 74.6 & 71.7 & 59.6 & 67.2 & 74.3 \\ 

\cmidrule{2-8}

& \multirow{4}{*}{\rotatebox[origin=c]{90}{\scriptsize{Weighted}}}

& 3 & 74.8 & 71.7 & 59.7 & 66.8 & 74.8\\

& & 7 & 74.9 & 72.1 & 60.3 & 67.3 & 74.8\\

& & 17 & 73.7 & 70.5 & 59.6 & 66.6 &  74.2\\

& & 31 & \textbf{75.1} & \textbf{72.3} & \textbf{61.0} & \textbf{67.6} &  \textbf{75.1}\\

\midrule


\multirow{8}{*}{\scriptsize{\textbf{G}}}

& \multirow{4}{*}{\rotatebox[origin=c]{90}{\scriptsize{Binary}}}

& 3  & 78.4 & 73.9 & 58.7 & 72.0 & 64.7\\

& & 7  & 79.2 & 75.4 & 59.4 & 72.8 & 65.2\\
 
& & 17 & 79.8 & 76.3 & 59.2 & 73.3 & 65.4\\

& & 31 & 79.3 & 75.7 & 57.5 & 73.4 & 65.5\\ 

\cmidrule{2-8}

&\multirow{4}{*}{\rotatebox[origin=c]{90}{\scriptsize{Weighted}}}

& 3 & 78.1 & 74.5 & 58.9 & 72.6 & 64.6\\

& & 7 & 79.6 & 75.9 & 60.1 & 73.0 & 65.5\\

& & 17 & 79.6 & 75.2 & 59.8 & 73.1 &  65.9\\

& & 31 & \textbf{81.5} & \textbf{77.5} & \textbf{60.8} & \textbf{75.6} &  \textbf{66.1}\\

\bottomrule

\end{tabular}
\caption{Ablation study on graph construction. Best results are in bold. \textbf{F}  denotes 50 Salads and \textbf{G} denotes GTEA.}
\label{tab:abl_graph_struct}
\end{table}

%% file: Tables/mlp_ablation.tex
\begin{table}[t]
\centering
\renewcommand{\arraystretch}{1.25}
\begin{tabular}{ c|c|c c c c c} 
\toprule
 & Method & F1@10 & F1@25 & F1@50 & Edit & Acc \\
\midrule

\multirow{2}{*}{\textbf{F}}

& MLP & 70.8 & 68.0 & 57.6 & 63.6 & 73.1 \\ 


& GCN & \textbf{75.1} & \textbf{72.3} & \textbf{61.0} & \textbf{67.6} & \textbf{75.1} \\ 

\midrule

\multirow{2}{*}{\textbf{G}}

& MLP & 77.8 & 74.4 & 59.1 & 72.2 & 65.6 \\ 


& GCN & \textbf{81.5} & \textbf{77.5} & \textbf{60.8} & \textbf{75.6} & \textbf{66.1} \\ 

\bottomrule
\end{tabular}
\caption{Ablation study on MLP vs. GCN. Best results are in bold. \textbf{F}  denotes 50 Salads and \textbf{G} denotes GTEA.}
\label{tab:mlp_ablation}
\end{table}

%% file: Tables/abl_alternating_learning.tex
\begin{table}[t]
\centering
\renewcommand{\arraystretch}{1.5}
\begin{tabular}{c|c|c|c c c c c}
\toprule
& $\gamma$ & $\epsilon$ & F1@10 &  F1@25 & F1@50 & Edit & Acc\\
\midrule

\multirow{4}{*}{\scriptsize{\textbf{F}}}

& 0 & 50 & 43.5 & 36.6 & 24.6 & 42.2 & 38.2\\

& 10 & 40 & 51.3 & 44.9 & 33.1 & 48.5 & 50.4\\

& 30 & 20 & \textbf{75.1} & \textbf{72.3} & \textbf{61.0} & \textbf{67.6} &  \textbf{75.1}\\

& 50 & 0 & 58.1 & 54.0 & 43.3 & 48.7 &  71.2\\

\midrule


\multirow{4}{*}{\scriptsize{\textbf{G}}}

& 0 & 50  & 21.6 & 20.3 & 15.9 & 16.9 & 29.6\\

& 10 & 40 & 22.8 & 21.7 & 16.9 & 17.3 & 29.9 \\

& 30 & 20 & \textbf{81.5} & \textbf{77.5} & \textbf{60.8} & \textbf{75.6} &  \textbf{66.1}\\

& 50 & 0 & 63.0 & 58.4 & 45.1 & 56.2 & 57.1\\ 

\bottomrule

\end{tabular}
\caption{Ablation study on alternating learning vs. joint learning. Best results are in bold. \textbf{F}  denotes 50 Salads and \textbf{G} denotes GTEA.}
\label{tab:abl_alternating_learning}
\end{table}

%% file: Tables/50sld_diff_suprvsn.tex
\begin{table}[t]
\centering
\renewcommand{\arraystretch}{1.5}
\begin{tabular}{c|c|c c c c c}
\toprule
& Method & F1@10 &  F1@25 & F1@50 & Edit & Acc\\
\midrule
\multirow{5}{*}{{\scriptsize{\textbf{F}}}}

& Baseline & 70.8 & 67.7 & 58.6 & 63.8 & 77.8 \\

& MS-TCN~\cite{farha2019ms} & 76.3 & 74.0 & 64.5 & 67.9 & 80.7\\

& MS-TCN++~\cite{li2020ms} & 80.7 & 78.5 & 70.1 & 74.3 & 83.7\\

& BCN~\cite{wang2020boundary} & 82.3 & 81.3 & 74.0 & 74.3 & 84.4\\

& ASRF~\cite{ishikawa2021alleviating} & \textbf{84.9} & \textbf{83.5} & \textbf{77.3} & \textbf{79.3} & \textbf{84.5} \\

\midrule

\multirow{2}{*}{{\scriptsize{\textbf{Ti}}}}

& ABE~\cite{li2021temporal} & 73.7 & 71.0 & 60.1 & 66.1 & \textbf{76.0} \\
 
& GCN (Ours) & \textbf{75.1} & \textbf{72.3} & \textbf{61.0} & \textbf{67.6} & 75.1\\

\midrule

\multirow{3}{*}{{\scriptsize{\textbf{Tr}}}}

& CDFL~\cite{li2019weakly} & - &  - & - & - & \textbf{54.7} \\

& NN-Viterbi~\cite{richard2018neuralnetwork} & - &  - & - & - & 49.4 \\

& HMM-RNN~\cite{richard2017weakly} & - &  - & - & - &  45.5 \\

\bottomrule

\end{tabular}
\caption{Comparison on 50 Salads. Best results are in bold. \textbf{F}  denotes Full Supervision, \textbf{Ti} denotes Timestamp Supervision, and \textbf{Tr} denotes Transcript Supervision.}
\label{tab:50salads_diff_supervision}
\end{table}

%% file: Tables/gtea_diff_suprvsn.tex
\begin{table}[t]
\centering
\renewcommand{\arraystretch}{1.5}
\begin{tabular}{c|c|c c c c c}
\toprule
& Method & F1@10 &  F1@25 & F1@50 & Edit & Acc\\
\midrule
\multirow{5}{*}{{\scriptsize{\textbf{F}}}}

& Baseline & 85.1 & 82.7 & 69.6 & 79.6 & 76.1 \\

& MS-TCN~\cite{farha2019ms} & 85.8 & 83.4 & 69.8 & 79.0 & 76.3\\

& MS-TCN++~\cite{li2020ms} & 88.8 & 85.7 & 76.0 & 83.5 & \textbf{80.1}\\

& BCN~\cite{wang2020boundary} &  88.5 & 87.1 & 77.3 & \textbf{84.4} & 79.8\\

& ASRF~\cite{ishikawa2021alleviating} & \textbf{89.4} & \textbf{87.8} & \textbf{79.8} & 83.7 & 77.3 \\

\midrule

\multirow{2}{*}{{\scriptsize{\textbf{Ti}}}}

& ABE~\cite{li2021temporal} & 77.7 & 73.8 & 58.1 & 72.1 & \textbf{67.7} \\
 
& GCN (Ours) & \textbf{81.5} & \textbf{77.5} & \textbf{60.8} & \textbf{75.6} & 66.1 \\

\bottomrule

\end{tabular}
\caption{Comparison on GTEA. Best results are in bold. \textbf{F}  denotes Full Supervision and \textbf{Ti} denotes Timestamp Supervision.}
\label{tab:gtea_diff_supervision}
\end{table}

%% file: Tables/breakfast_diff_suprvsn.tex
\begin{table}[t]
\centering
\renewcommand{\arraystretch}{1.5}
\begin{tabular}{c|c|c c c c c}
\toprule
& Method & F1@10 &  F1@25 & F1@50 & Edit & Acc\\
\midrule
\multirow{5}{*}{{\scriptsize{\textbf{F}}}}

& Baseline & 69.9 & 64.2 & 51.5 & 69.4 & 68.0 \\

& MS-TCN~\cite{farha2019ms} & 52.6 & 48.1 & 37.9 & 61.7 & 66.3\\

& MS-TCN++~\cite{li2020ms} & 64.1 & 58.6 & 45.9 & 65.6 & 67.6\\

& BCN~\cite{wang2020boundary} & 68.7 & 65.5 & 55.0 & 66.2 & \textbf{70.4} \\

& ASRF~\cite{ishikawa2021alleviating} &  \textbf{74.3} & \textbf{68.9} & \textbf{56.1} & \textbf{72.4} & 67.6 \\

\midrule

\multirow{2}{*}{{\scriptsize{\textbf{Ti}}}}

& ABE~\cite{li2021temporal} & 67.4 & 60.8 & 44.9 & \textbf{68.5} & \textbf{63.1} \\
 
& GCN (Ours) & \textbf{67.9} & \textbf{61.0} & \textbf{45.3} & 67.0 & 61.4\\

\midrule

\multirow{7}{*}{{\scriptsize{\textbf{Tr}}}}

& CDFL~\cite{li2019weakly} & - &  - & - & - & \textbf{50.2} \\

& MuCon\cite{souri2021fast} & - &  - & - & - & 47.1 \\

& \(D^3\)TW~\cite{chang2019d3tw} & - &  - & - & - &   45.7 \\

& NN-Viterbi~\cite{richard2018neuralnetwork}  & - &  - & - & - &  43.0 \\

& TCFPN~\cite{ding2018weakly} & - &  - & - & - &   38.4 \\

& HMM-RNN~\cite{richard2017weakly} & - &  - & - & - &  33.3 \\

& ECTC~\cite{huang2016connectionist} & - &  - & - & - &  27.7 \\

\midrule

\multirow{3}{*}{{\scriptsize{\textbf{S}}}}

&  SCT~\cite{fayyaz2020sct} & - &  - & - & - &  \textbf{30.4} \\

& SCV~\cite{li2020set} & - &  - & - & - &   30.2 \\

& Action Sets~\cite{richard2018action} & - &  - & - & - &  23.3 \\

\bottomrule

\end{tabular}

\caption{Comparison on Breakfast. Best results are in bold. \textbf{F}  denotes Full Supervision, \textbf{Ti} denotes Timestamp Supervision, \textbf{Tr} denotes Transcript Supervision, and \textbf{S} denotes Set Supervision.}
\label{tab:breakfast_diff_supervision}
\end{table}

%% file: Tables/desktopassembly_diff_suprvsn.tex
\begin{table}[t]
\centering
\renewcommand{\arraystretch}{1.5}
\begin{tabular}{c|c|c c c c c}
\toprule
& Method & F1@10 &  F1@25 & F1@50 & Edit & Acc\\
\midrule
\multirow{2}{*}{{\scriptsize{\textbf{F}}}}

& Baseline & 90.2 & 87.2 & 76.9 & \textbf{89.9} & 79.4 \\

& ASRF~\cite{ishikawa2021alleviating} & \textbf{91.4} & \textbf{90.3} & \textbf{83.2} & 86.6 & \textbf{81.9} \\ 

\midrule

\multirow{2}{*}{{\scriptsize{\textbf{Ti}}}}

& ABE~\cite{li2021temporal} & 89.8 & 86.4 & 67.5 & \textbf{88.3} & 71.7 \\ 

& GCN (Ours) & \textbf{90.4} & \textbf{88.0} & \textbf{75.1} & 87.3 & \textbf{77.1} \\ 

\bottomrule

\end{tabular}

\caption{Comparison on Desktop Assembly. Best results are in bold. \textbf{F} denotes Full Supervision and \textbf{Ti} denotes Timestamp Supervision.}
\label{tab:desktopassembly_diff_supervision}
\end{table}

%% file: Sections/conclusion.tex
\section{Conclusion}
\label{sec:conclusion}

We propose, in this paper, a novel method for timestamp-supervised activity segmentation, which utilizes a graph convolutional network for generating dense framewise labels from sparse timestamp labels. The graph convolutional network is learned in an end-to-end fashion to leverage not only frame features but also connections between neighboring frames. Moreover, we present a framework for alternating learning of both the segmentation model and the graph convolutional model. We show that our method is superior to the multi-layer perceptron baseline, while performing on par with or better than the state of the art in timestamp-supervised activity segmentation on 50 Salads, GTEA, Breakfast, and Desktop Assembly. Our future work will explore the use of deep supervision~\cite{lee2015deeply,li2017deep,li2018deep,fathy2018hierarchical,zhuang2019degeneracy} or self-supervised losses~\cite{mobahi2009deep,haresh2021learning} for improving the performance.